\pdfoutput=1

\documentclass[11pt]{article}
\usepackage[]{emnlp2021}
\usepackage{times}
\usepackage{latexsym}
\usepackage[T1]{fontenc}
\usepackage[utf8]{inputenc}
\usepackage{microtype}
\usepackage{booktabs}

\usepackage[noend]{algpseudocode}
\usepackage{algorithmicx,algorithm}
\usepackage{multirow}
\usepackage{amsmath}
\DeclareMathOperator*{\argmax}{argmax}

\newcommand*\samethanks[1][\value{footnote}]{\footnotemark[#1]}

\author{Xingran Chen\textsuperscript{3}\thanks{\quad The two authors contributed equally to this work.}\space, Ge Zhang\textsuperscript{1\space 2\space 3}\samethanks\space, Adam Nik\textsuperscript{2\space 4}, Mingyu Li\textsuperscript{2\space 3},  Jie Fu\thanks{\quad Corresponding Author}\; \textsuperscript{1}\\ \textsuperscript{1} Beijing Academy of Artificial Intelligence, China\\ 
\textsuperscript{2} 1Cademy Community, USA\\  
\textsuperscript{3} University of Michigan Ann Arbor, USA\\
\textsuperscript{4} Carleton College, USA\\
\texttt{\href{fujie@baai.ac.cn}{\color{black}{fujie AT baai.ac.cn}}}
}

\title{1Cademy @ Causal News Corpus 2022: Enhance Causal Span Detection via Beam-Search-based Position Selector}

\begin{document}
\maketitle
\begin{abstract}
In this paper, we present our approach and empirical observations for Cause-Effect Signal Span Detection---Subtask 2 of Shared task 3~\cite{tan-etal-2022-event} at CASE 2022. 
The shared task aims to extract the cause, effect, and signal spans from a given causal sentence.
We model the task as a reading comprehension (RC) problem and apply a token-level RC-based span prediction paradigm to the task as the baseline.
We explore different training objectives to fine-tune the model, as well as data augmentation (DA) tricks based on the language model (LM) for performance improvement.
Additionally, we propose an efficient beam-search post-processing strategy to due with the drawbacks of span detection to obtain a further performance gain.
Our approach achieves an average $F_1$ score of 54.15 and ranks \textbf{$1^{st}$} in the CASE competition. Our code is available at \url{https://github.com/Gzhang-umich/1CademyTeamOfCASE}.

\end{abstract}

\section{Introduction}

 
 Event extraction has long been a challenging and popular area for natural language processing (NLP) researchers. 
 There are known classic benchmarks, including ACE-2005~\cite{ACE_2005} and ERE~\cite{song-etal-2015-light}.
 In recent years, more and more interesting corpora about event detection and extraction have emerged based on different specific source corpora, including biomedical literature~\cite{kim2003genia}, scientific knowledge resources~\cite{jain-etal-2020-scirex}, Wiki~\cite{li-etal-2021-document}, and trade-related news~\cite{zhou-etal-2021-trade}.   
 In sharp contrast, Cause-Effect Signal Span Detection aims to extract the cause, effect, and signal spans from sentences that have cause-effect relations. 
 Cause-Effect Signal Span Detection is an innovative and important event detection/extraction task that assists in understanding causal relationships from comprehensive sentence samples.
 
 As a new corpus with great potential in event extraction challenges, the Causal News Corpus (CNC) \cite{CNC_Tan_2022} contains socio-political event (SPE) text data with annotated causal spans. 
 The CNC event extraction challenge\footnote{\url{https://github.com/tanfiona/CausalNewsCorpus}} is the first Cause-Effect Signal Span Detection challenge on a social political news corpus.
The challenge itself provides a limited number of annotated samples for supervision, making it more difficult compared to other challenging event extraction tasks.
 The exploration of causality in news data and the detection of corresponding spans is helpful in reading comprehensive language expressions, making CNC attractive to NLP researchers.

In this paper, we describe our RC-based model with a carefully designed post-processing strategy. 
We also conduct ablation studies to analyze the influence of both different training objectives and different hyper-parameter settings of the post-processing strategy on our model.
In addition, we apply an LM-based data augmentation strategy to further better performance gains, given the low-resource challenge. 
Our approach improves performance by a large margin in Cause-Effect Signal Span Detection compared to any other competitors.

The main contributions of our paper are as follow:
\begin{itemize}
    \item We propose an RC-based model with an original post-processing strategy.
    \item We achieve state-of-the-art performance on the new Cause-Effect Signal Span Detection competition on the CNC.
    \item We apply an LM-based data augmentation technique to the challenge and prove its positive effect on the challenge of low resources.
\end{itemize}

\section{Causal News Corpus}

The corpus we used in our model training and evaluation is the CNC dataset \citep{CNC_Tan_2022}. 
This dataset is built on the extraction of social-political events from News (AESPEN)~\citep{hurriyetoglu-etal-2020-automated} in 2020  and the CASE 2021 workshop @ ACL-IJCNLP \citep{hurriyetoglu-etal-2021-challenges}. Each sample in the dataset is annotated with causal labels, that is, whether a sentence contains a causal event. Furthermore, some sentences are annotated with the span of the specific Cause and Effect of a causal event, as well as the signal markers that imply the causality. The spans are labeled by <ARG0>, <ARG1>, and <SIG> annotations to represent the cause, effect, and causal signal in the sentence, respectively. Note that it is possible to have multiple annotations for the same sentence in the dataset if the sentence contains multiple casual relationships of events. The dataset statistics are shown in Table~\ref{tab:statistics}.

\begin{table}[!t]
\centering
\caption{Dataset statistics. Avg. Singal represents the average number of Signal spans in each split of dataset.}
\label{tab:statistics}
\begin{tabular}{lllll}
\toprule
             & Train & Valid & Test & Total \\ \midrule
\# Sentences & 160   & 15    & 89   & 264   \\
\# Relations & 183   & 18    & 119  & 320   \\
Avg. Signal  & 0.67  & 0.56  & 0.82 & 0.72  \\ \bottomrule
\end{tabular}
\end{table}


\section{Methodology}

In this section, we describe in detail the methodology we used in the task. To begin, we introduce the baseline model established from a pre-trained language model for the task. Next, a beam-search-based post-processing method is introduced to solve the overlap span detection problem in the baseline model. 
To address the problem that not all examples have signal markers within the sentence, we propose training a signal classifier to determine whether we need to find the signal span of the target test sample. 
Finally, a pre-trained paraphrasing model is applied for data augmentation.
\begin{algorithm}[t]
\small
\caption{beam-search-based span selector}
\label{algo-main}
\hspace*{0.02in} {\bf Input:}
$P_{s_c}, P_{e_c}, P_{s_{ef}}, P_{e_{ef}}, n, k, m$. \\
\hspace*{0.02in} {\bf Output:} 
$H = \{(s_1, e_1, s_2, e_2, t_i=CBeforeE/CAfterE) : i \leq m \}$ 
\begin{algorithmic}[1]
\State CBeforeE = $\{p^i_{s_c} + p^j_{e_{ef}}: 1 \leq i,j \leq n\}$.
\State CAfterE = $\{p^i_{s_{ef}} + p^j_{e_c}: 1 \leq i,j \leq n\}$.
\State Find position pairs with Top-$k$ largest score from both CBeforeE and CAfterE. 
\State Denote the gotten position pairs set as $PS$ = $\{(sp_i,ep_i,t_i = CBeforeE/CAfterE) : sp_i \leq ep_i \}$. $t_i$ implies whether the pair is retrieved from CBeforeE or CAfterE. 
\State Initialize a min heap $H$.
\For {$ps_p = (sp_p, ep_p, t_p)$ in $PS$} 
    \If {$t_p$ = $CBeforeE$} 
        \State Find the position pair $(i,j)$ with the largest $p^i_{e_c} + p^j_{s_{ef}}$, which satisfies $sp_p \leq i \leq j \leq ep_p$.
        \State Calculate $sc_{(sp_p,i,j,ep_p)} = p^{sp_p}_{s_c} + p^i_{e_c} + p^j_{s_{ef}} + p^{ep_p}_{e_{ef}}$.
    \Else
        \State Find the position pair $(i,j)$ with the largest $p^i_{e_{ef}} + p^j_{s_{c}}$, which satisfies $sp_p \leq i \leq j \leq ep_p$.
        \State Calculate $sc_{(sp_p,i,j,ep_p)} = p^{sp_p}_{s_{ef}} + p^i_{e_{ef}} + p^j_{s_{c}} + p^{ep_p}_{e_{c}}$.
    \EndIf
    \State Push $\{(sp_p,i,j,ep_p),t_p, sc_{(sp_p,i,j,ep_p)}\}$ into $H$.
    \If {$len(H) > m$}
        \State $heappop(H)$ based on $sc_{(sp_p,i,j,ep_p)}$.
    \EndIf
\EndFor
\State \Return $H$
\end{algorithmic}
\end{algorithm}

\subsection{Baseline}
\label{baseline}

To solve the task, we first fine-tune the pre-trained language model based on the reading comprehension training fashion proposed by BERT~\cite{devlin-etal-2019-bert}. Specifically, assume that we need to predict a span within sentence $x = \{t_1, ... , t_n\}$, where $t_i$ is the \textit{i}$^{th}$ token of sentence $x$. We can obtain a contextualized representation $h_i$ of $t_i$ using the pre-trained language model:

\begin{equation}
    H = \{h_1, ..., h_n\} = BERT(x)
\end{equation}

Next, we define two parameterized vectors: $v_s, v_e \in R^d$ to calculate the probability that the \textit{i}$^{th}$ token is the start / end position:

\begin{equation}
    P_s = \{p_s^{(1)},...,p_s^{(n)}\} = Softmax(v_s^TH)
\end{equation}

\begin{equation}
    P_e = \{p_e^{(1)},...,p_e^{(n)}\} = Softmax(v_e^TH)
\end{equation}

We select the positions with maximum probability as the prediction of the model:

\begin{equation}
    s = \argmax_{1 \leq i \leq n} p_s^{(i)},
\end{equation}

\begin{equation}
    e = \argmax_{1 \leq j \leq n} p_e^{(j)},
\end{equation}
where $s, e$ represent the predicted start/end position, respectively.

The prediction of the spans of cause, effect, and signal are all similar to the span prediction task described above. For convenience, we will denote the start/end position of cause, effect, and signal as $s_{c}, e_{c}, s_{ef}, e_{ef}, s_{sig}, e_{sig}$, respectively, to specify which span we are detecting. Therefore, the training objective is to maximize the probability of ground-truth positions in the model.

\subsection{Beam-search-based Span Selector}

The proposed baseline model has two drawbacks. First, it is possible that the end position is right before the start position. Second, it is possible to generate spans that overlap each other, which is not allowed in the challenge. 
Thus, we need to introduce constraints in post-processing to ensure that: 1) the predicted end position must be after the start position of the same span, and 2) the predicted spans of cause and effect do not overlap with each other. 
In this sub-section, we describe our modified beam search-based algorithm to address the overlapping issue. 
The beam search algorithm is widely used to find the most possible output with tractable memory and time usage in text generation tasks~\cite{xie2017neural}. 
In reading comprehension or question answering, it is also used to introduce constraint information~\cite{hu2019multi}, and therefore encourage more accurate predictions.
Given a paragraph with length $n$, we can calculate $P_{s_c} = \{p^{(1)}_{s_c}, ... ,p^{(n)}_{s_c}\}$ based on the process introduced in \S~\ref{baseline}.
Similarly, we can calculate $P_{e_c}$, $P_{s_{ef}}$, and $P_{e_{ef}}$ accordingly.
Formally, given the input probability vectors $P_{s_c}$, $P_{e_c}$, $P_{s_{ef}}$, $P_{e_{ef}}$, a hyper-parameter $m$ denoting the requested answer number, and a hyper-parameter $k$ denoting the beam search size, the span selector is expected to output position pairs $s_c$, $e_c$, $s_{ef}$ and $e_{ef}$.
We describe the span selector in detail in Algorithm~\ref{algo-main}.
We denote the proposed span selector as \textbf{BSS}. It should be noted that the proposed BSS post-processing algorithm can also generate multiple predictions for cases containing multiple causal relations. For example, we could change the hyperparameter $m$ to retrieve the prediction of cause/effect spans combinations with the top-$m$ highest scores as our predictions of multiple causal relations. For the signal span, we always use the span with the highest score as our prediction (if it presents).

\subsection{Signal Classifier}
\label{sec:signal_classifier}

We observe that some samples do not have signal markers (spans) within the sentence even while the baseline model predicts $s_{sig}, e_{sig}$ for each target sample. Therefore, we propose to train a classifier to address this issue. Specifically, we first automatically annotate training samples based on whether signal markers appear within the samples. Then, we fine-tune the pre-trained language model to train a binary classifier. Note that we can share the language model parameters between signal classifier and span detection, i.e. we optimize both training objectives during our fine-tuning process. In addition, we can also train a signal classifier with a separate language model. In our experiments, we apply the two methods separately and compare their effectiveness.

\begin{table*}[ht]
\centering
\caption{Experimental results and related ablation study on subtask 2. The evaluation metric of all the results is $F_1$. Note that $n$ represents the hyper-parameter of data augmentation described in \S~\ref{sec:da}.}
\begin{tabular}{lcccc}
\toprule
\textbf{Methods}     & \textbf{Cause} & \textbf{Effect} & \textbf{Signal} & \textbf{Overall} \\ \midrule
Baseline                                    & 77.8  & 66.7   & 53.5   & 68.2    \\
Baseline-NER                              & 57.8  & 57.4   & 10.8   & 47.4    \\ \midrule
Baseline + DA ($n = 2$)                     & 72.2  & 77.8   & 60.9.  & 71.9    \\
Baseline + BSS + DA ($n = 2$)               & 77.8     & \textbf{83.3}      & 60.9      & 74.1       \\
Baseline + ES  + DA ($n = 2$)               & 72.2     & 77.8      & 76.7      & 75.4       \\
Baseline + JS + DA ($n = 2$)                & 72.2     & 72.2      & 71.3      & 69.8       \\
Baseline + BSS + ES + DA ($n = 2$)          & 77.8     & \textbf{83.3}      & 76.7      & 77.5       \\
Baseline + BSS + ES + DA ($n = 3$)          & \textbf{83.3}     & 77.8      & \textbf{80.0}      & \textbf{80.4}       \\ \bottomrule
\end{tabular}
\label{tab:main}
\end{table*}

\subsection{Data Augmentation with Pre-trained Paraphrasing Model}
\label{sec:da}
Considering that only 183 training samples are available for subtask 2, it is important to introduce the data augmentation trick to increase the size of the training dataset. Therefore, in this work, we propose using language models to paraphrase the existing data. Specifically, we use a PEGASUS model~\cite{zhang2020pegasus} fine-tuned for paraphrasing~\footnote{We directly use fine-tuned checkpoint in https://huggingface.co/tuner007/pegasus\_paraphrase} to re-write the phrases of Cause, Effect in each sample. 
For example, for a training sample "\textit{<ARG1>The farmworkers ' strike resumed on Tuesday</ARG1> when <ARG0>their demands were not met</ARG0>.}", we paraphrase the cause and effect spans within the sample, then obtain the augmented sample "\textit{<ARG1>On Tuesday, the farmworkers resumed their strike</ARG1> when <ARG0>their demands weren't met</ARG0>.}". In this case, the semantic meaning of the original sentence is preserved. Hence, the annotation of the original sample is still reasonable and can continue to be used in the augmented sample. In our implementation, $n$ new phrases were generated for each span. Namely that each sample will end up with $n^2$ augmented samples. We denote the trick as {\bf DA}.

\section{Experiments}
In this section, we present the experimental details of training the model and discuss the performance of our proposed approach.

\subsection{Experimental Details}
In our experiment, we use Albert~\cite{lan2019albert} as our LM backbone. 
We perform hyper-parameter searching to find the best hyper-parameter setting. Specifically, we select the learning rate $l$ from $\{1e-5, 2e-5, 5e-5\}$, batch size $b$ from $\{1, 2, 4, 8, 16, 32\}$. We fine-tune the pre-trained model for 30 epochs, and select the checkpoint with the best performance on the development set to conduct evaluation on the test set. Our implementation is based on \texttt{Huggingface}~\cite{wolf2019huggingface}.

In terms of the signal classifier, we consider two settings: 1) We fine-tune the signal classifier in conjunction with the main training objective as described in \S~\ref{sec:signal_classifier}. We denote this approach as \textbf{Joint Sig. (JS)}; 2) We additionally fine-tune a language model to specifically decide whether to predict the span of Signal. We denote this approach by \textbf{Extra Sig. (ES)}

We also include another implementation of the baseline recommended by the organizers, where the fine-tuning process is carried out in the end-to-end fashion of Named Entity Recognition (NER). We denote this baseline by \textbf{Baseline-NER}.

\subsection{Main Results and Ablation Study}

Here, we present and discuss the experimental results of our best-performing method for this task, together with the corresponding ablation study. 
Note that all results are evaluated on the dev set, due to the inaccessibility of the test dataset. 
We present the score of different approaches $F_1$ on all three span detection in Table~\ref{tab:main}.

The results clearly show that the reading comprehension style of the training significantly improves the effectiveness of the approach. 
We can also observe that it is better to apply the reading comprehension training fashion than token-level tagging for the causal span detection task. 
Regarding our proposed approaches, the LM-based paraphrasing data augmentation technique improves the performance of the approach by a large margin compared to the baseline. The improvement is consistent, that is, there is an improvement in the prediction of all types of spans. In addition, our proposed BSS post-processing algorithm further improves our approach. However, it can be seen that the improvement of the approach by BSS mainly comes from the prediction of cause and effect. This is reasonable because the algorithm does not post-process the predictions of Signal. As for the signal classifier, both ES and JS make an improvement, which comes mainly from the better prediction of Signal. However, note that the improvement in ES is larger.
We conjecture that it might be because of a new training objective introduced by JS, which is harmful to the proposed approach to learning to predict the spans better. 
Finally, we mix all of the approaches together with our approach and ended up with the best performance. Here, we also compared the impact of data augmentation at different scales. Specifically, we compare the results when $n = 2$ ($4 \times$ dataset size) with $n = 3$ ($9 \times$ dataset size). We find that higher data augmentation sizes lead to better results in the validation dataset.

\subsection{Case Study of Data Augmentation}

\begin{table*}[!h]
\centering
\small
\caption{Case Study of Data Augmentation. Note that we generate two sentences for Cause and Effect, respectively. Therefore, there are in total 4 outcomes sentences via combinations.}
\label{tab:casestudy}
\begin{tabular}{ll} \toprule
Ori. &
  \textless{}ARG1\textgreater{}The farmworkers ' strike resumed on Tuesday\textless{}/ARG1\textgreater when \textless{}ARG0\textgreater{}their demands were not met\textless{}/ARG0\textgreater{} \\ \midrule
\multirow{4}{*}{DA} &
  \textless{}ARG1\textgreater{}On Tuesday, the farmworkers resumed their strike\textless{}/ARG1\textgreater when \textless{}ARG0\textgreater{}their demands weren't met\textless{}/ARG0\textgreater . \\
 &
  \textless{}ARG1\textgreater{}On Tuesday, the farmworkers resumed their strike\textless{}/ARG1\textgreater when \textless{}ARG0\textgreater{}their demands didn't get met\textless{}/ARG0\textgreater . \\
 &
  \textless{}ARG1\textgreater{}On Tuesday, the farmworkers went on strike\textless{}/ARG1\textgreater when \textless{}ARG0\textgreater{}their demands weren't met\textless{}/ARG0\textgreater . \\
 &
  \textless{}ARG1\textgreater{}On Tuesday, the farmworkers went on strike\textless{}/ARG1\textgreater when \textless{}ARG0\textgreater{}their demands didn't get met\textless{}/ARG0\textgreater . \\ \bottomrule
\end{tabular}
\end{table*}

In this subsection, we provide a case study on the effectiveness of data augmentation proposed in the system. The comparisons between generated texts and the original texts are shown in Table~\ref{tab:casestudy}. 

From the results, the expressions in the data-augmented texts are more diverse while remaining semantically consistent with the original sentence. Furthermore, the data-augmented texts are competitive with the original in terms of fluency and grammatical correctness.

\subsection{Competition Result}
We reveal and discuss the final results of our proposed approach competition on a test set. The results are shown in Table~\ref{tab:contest}.

As shown in the table, our proposed approach achieves state-of-the-art results in 3 out of 4 evaluation metrics on subtask 2. This shows the excellent performance of the proposed approach in solving the task of causal spans detection.

\begin{table}[]
\label{sec:submission}
\caption{Overall performance of the proposed approach on the test set. The numbers in parentheses represent the rankings.}
\centering
\begin{tabular}{lr}
\toprule
\multicolumn{2}{c}{Final Competition Results} \\ \midrule
\multicolumn{1}{l|}{Recall}         & 0.5387 (1)   \\
\multicolumn{1}{l|}{Precision}      & 0.5509 (2)   \\
\multicolumn{1}{l|}{F1}             & 0.5415 (1)   \\
\multicolumn{1}{l|}{Accuracy}       & 0.4315 (1)   \\ \bottomrule
\end{tabular}
\captionsetup{justification=centering}
\label{tab:contest}
\end{table}









\section{Conclusion}

 This paper introduces a reading comprehension-based method, an original post-processing strategy, and an LM-based data augmentation trick for the new Cause-Effect Signal Span Detection competition. 
 We compare the RC-based method with the NER-based one and prove that the RC-based method gets an observing performance gain compared to the NER-based one. 
 We provide experimental results and ablation studies of our beam-search-based Span Selector and LM-based data augmentation tricks to analyze their efficiency and prove their compatibility with other tricks.
 Our approach achieves state-of-the-art performance in the new competition.

\section*{Acknowledgements}

This study was supported by the National Key R\&D Program of China (2020AAA0105200).

\bibliography{anthology,custom}
\bibliographystyle{acl_natbib}

\end{document}